\documentclass{article}
\usepackage{iclr2027_conference,times}
\iclrfinalcopy

\usepackage[utf8]{inputenc}
\usepackage[T1]{fontenc}

\usepackage{hyperref}
\usepackage{url}

\usepackage{booktabs}
\usepackage{multirow}

\usepackage{amsmath}
\usepackage{amsfonts}
\usepackage{amssymb}
\usepackage{nicefrac}

\usepackage{graphicx}
\usepackage{caption}
\usepackage{subcaption}

\usepackage{algorithm}
\usepackage{algpseudocode}
\usepackage{float}
\usepackage{placeins}

\usepackage{microtype}
\usepackage{xcolor}

\newcommand{\ours}{\textsc{POMA}}
\newcommand{\bench}{\textsc{SCOPE-Bench}}
\definecolor{bestcolor}{HTML}{0F6E56}
\title{Learning to Select Source Domains:\\Proxy-Rewarded Policy Optimization\\for Molecular OOD Generalization}

\author{%
  \begin{minipage}[t]{0.96\textwidth}
  \raggedright\normalfont
  \textbf{Zhuohao Lin, Kun Li, Jiameng Chen, Wenbin Hu$^{*}$}\\[3pt]
  \small School of Computer Science, Wuhan University\\
  Wuhan, China\\
  \texttt{\{linzhuohao, likun98, jiameng.chen, hwb\}@whu.edu.cn}\\[3pt]
  $^{*}$Corresponding author.
  \end{minipage}
  \AND
  \begin{minipage}[t]{0.29\textwidth}
  \raggedright\normalfont\small
  \textbf{Yizhen Zheng}\\[3pt]
  Department of Data Science and Artificial Intelligence,\\
  Monash University\\
  Victoria, Australia\\
  {\fontsize{8}{9}\selectfont\texttt{yizhen.zheng1@monash.edu}}
  \end{minipage}
  \And
  \begin{minipage}[t]{0.29\textwidth}
  \raggedright\normalfont\small
  \textbf{Jiajun Yu}\\[3pt]
  College of Computer Science and Technology,\\
  Zhejiang University\\
  Hangzhou, China\\
  {\fontsize{8}{9}\selectfont\texttt{jiajunyu1999@gmail.com}}
  \end{minipage}
  \And
  \begin{minipage}[t]{0.29\textwidth}
  \raggedright\normalfont\small
  \textbf{Duanhua Cao}\\[3pt]
  School of Life Sciences and Technology,\\
  Tongji University\\
  Shanghai, 200092, China\\
  {\fontsize{8}{9}\selectfont\texttt{caodh@tongji.edu.cn}}
  \end{minipage}
}

\hypersetup{hidelinks,
pdfauthor={Zhuohao Lin, Kun Li, Jiameng Chen, Wenbin Hu, Yizhen Zheng, Jiajun Yu, Duanhua Cao}, pdftitle={Learning to Select Source Domains: Proxy-Rewarded Policy Optimization for Molecular OOD Generalization} }

\begin{document}

\maketitle
\fancyhead{}
\fancyhead[L]{\small Preprint}
\suppressfloats[t]

\begin{abstract}
Robust molecular property prediction under extreme out-of-distribution (OOD) shifts faces two challenges. Conventional scaffold splits can retain local structural similarity across training and test molecules, while adaptation from heterogeneous source domains can lead to negative transfer. We introduce Policy Optimization for Multi-source Adaptation (POMA), a retrieve--compose--adapt framework for adaptive source domain selection when target property labels are unavailable. POMA uses unlabeled target structures to retrieve structurally similar labeled source scaffolds as proxy targets. A reinforcement learning policy learns to explore combinations of candidate source domains, using post-adaptation reductions in proxy-task mean absolute error (MAE) as rewards. The predictor is then adapted using the selected sources, with covariance alignment at the whole-molecule and BRICS-derived substructure levels. We also introduce SCOPE-Bench, a scaffold-cluster benchmark constructed in an explicit physicochemical descriptor space. Experiments on QM9 evaluate HOMO, LUMO, and HOMO--LUMO gap prediction across three 3D molecular backbones and 15 target-scaffold tasks. Prediction errors on SCOPE-Bench increase by up to $8.0\times$, with a mean of $5.9\times$, relative to conventional scaffold splitting. Relative to the supervised baseline, POMA reduces MAE by up to 11.2\%, averaging 6.2\% over the nine evaluated backbone--property combinations. Code is available at \url{https://anonymous.4open.science/r/Molecular-OOD-Code-73F6}.

\end{abstract}

\section{Introduction}
\label{sec:intro}

Recent 3D geometric models, including ViSNet, ETNN, and GotenNet, encode molecular geometry in learned representations and achieve strong performance on standard molecular benchmarks~\citep{visnet,etnn,gotennet}. Yet benchmark accuracy alone does not establish reliability under structural distribution shifts~\citep{ood_survey_shen,ood_graph_survey}. We study how to evaluate molecular out-of-distribution (OOD) generalization beyond conventional scaffold splits and how to select source subsets when target property labels are unavailable.

\begin{figure}[t]
  \centering
  \includegraphics[width=0.9\linewidth]{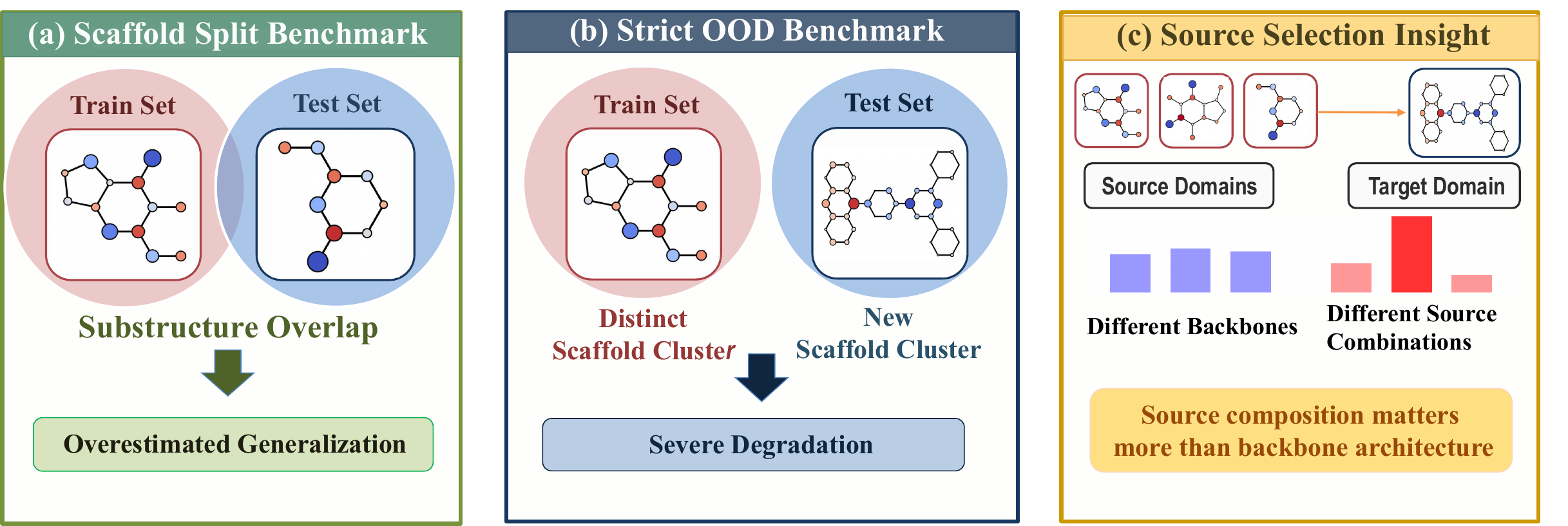}
\caption{Conceptual motivation for molecular OOD evaluation and source selection. (a) Distinct scaffolds can share local substructures. (b) Descriptor-based scaffold clustering defines source and target partitions. (c) Source composition is an adaptation decision. All panels are schematic; panel (c) does not quantify the relative effects of source composition and backbone choice.}
  \label{fig:motivation}
\end{figure}

Scaffold splitting groups molecules by their Bemis--Murcko scaffolds and assigns scaffold groups to disjoint data splits~\citep{bemis_murcko,moleculenet}. However, distinct scaffolds may still share local substructures relevant to property prediction, so scaffold disjointness alone does not guarantee separation of these patterns. We introduce \bench{}, a scaffold-cluster OOD benchmark that clusters scaffolds in an explicit physicochemical descriptor space and partitions the resulting clusters into source, validation, and target domains~\citep{moleood}. This descriptor-based cluster split complements conventional scaffold splitting by evaluating prediction under cluster-level structural shifts.

Beyond evaluation, adaptation requires deciding which source domains to use. Transfer from heterogeneous labeled sources can be beneficial~\citep{transfer_learning_survey,m3sda}, but poorly matched sources may induce negative transfer~\citep{negative_transfer}. Our ViSNet--HOMO comparisons illustrate this sensitivity. With the dual-scale adaptation design fixed across selection strategies, some strategies improve over the supervised baseline whereas others perform worse (Table~\ref{tab:ablation}). We therefore treat source subset selection as an explicit adaptation decision. Structural similarity can retrieve candidates but does not directly measure the downstream benefit of adapting with a source subset. Learning this benefit without target property labels therefore requires a surrogate feedback signal. Figure~\ref{fig:motivation} illustrates these evaluation and source-selection motivations.

We propose Policy Optimization for Multi-source Adaptation (\ours{}), a retrieve--compose--adapt framework that learns source-subset selection from proxy-task adaptation feedback. Unlabeled target structures guide the retrieval of structurally similar scaffolds from labeled source domains as proxy targets. \ours{} evaluates sampled source subsets through adaptation to these proxy targets and uses post-adaptation reductions in proxy-task mean absolute error (MAE) as rewards. These rewards train a reinforcement learning policy over the combinatorial source-subset space using Group Relative Policy Optimization~\citep{deepseekmath}. Structural similarity thus guides retrieval, while proxy-task adaptation performance supplies the learning signal for source selection.

\begin{figure}[ht!]
  \centering
  \includegraphics[width=0.88\linewidth]{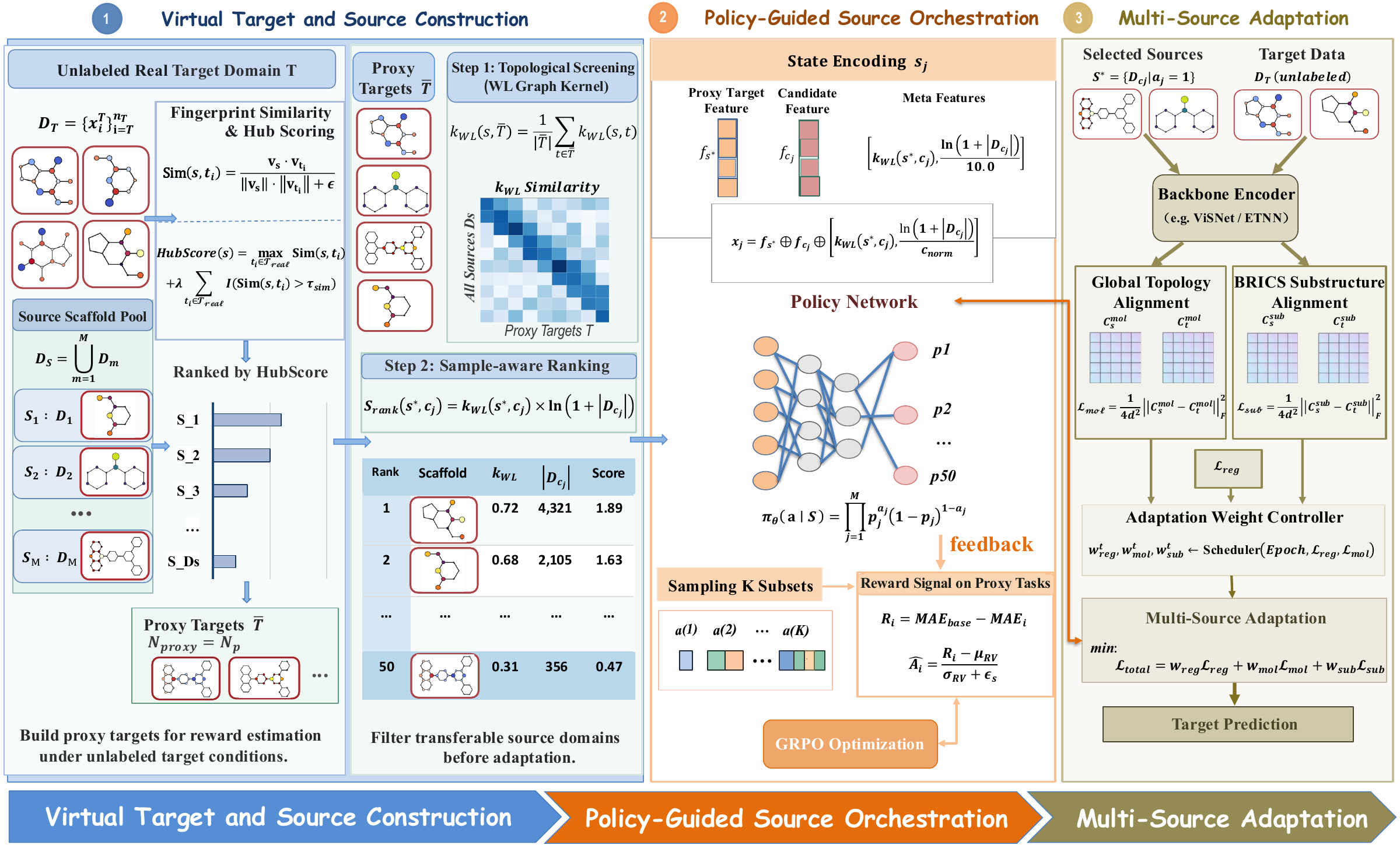}
\caption{Overview of \ours{}. Unlabeled target structures guide the retrieval of scaffolds from labeled source domains as proxy targets. Sampled source subsets are evaluated through dual-scale adaptation on proxy tasks, and the resulting MAE reductions provide policy rewards. The trained policy selects a source subset for the real target domain; the predictor is then adapted using labeled molecules from the selected source domains and unlabeled real-target structures, with covariance alignment at the whole-molecule and BRICS-derived substructure levels.}
  \label{fig:framework}
\end{figure}

The selection policy is coupled with a dual-scale adaptation module. For each selected subset, \ours{} combines supervised regression on labeled source molecules with covariance alignment between source and target representations at the whole-molecule and BRICS-derived substructure levels~\citep{brics,deep_coral}. The same adaptation procedure is used both to evaluate source subsets on proxy tasks and to adapt the predictor to the real target domain using labeled molecules from the selected sources and unlabeled target structures. Policy feedback is therefore tied to the downstream adaptation procedure, as summarized in Figure~\ref{fig:framework}.

Our contributions are threefold:

\textbf{(i) Scaffold-cluster OOD evaluation.} We introduce \bench{} to evaluate molecular prediction under descriptor-based scaffold-cluster shifts. Across the nine evaluated backbone--property combinations on QM9, the MAE under \bench{} is as much as $8.0\times$ the corresponding MAE under conventional scaffold splitting; the mean degradation factor is $5.9\times$.

\textbf{(ii) Proxy-reward-driven source selection.} We develop \ours{}, which learns source-subset selection from proxy-task adaptation rewards and couples the learned policy with covariance alignment at the whole-molecule and BRICS-derived substructure levels.

\textbf{(iii) Empirical evaluation.} Experiments on QM9 cover HOMO, LUMO, and HOMO--LUMO gap prediction across three 3D molecular backbones and 15 target-scaffold tasks. Relative to the supervised baseline, \ours{} reduces MAE by up to 11.2\%, with an average relative MAE reduction of 6.2\% over the nine evaluated backbone--property combinations. Source-selection comparisons and alignment ablations examine the framework components within this setting.

\section{Related Work}
\label{sec:related}

\paragraph{Molecular representation learning.} Molecular representations range from predefined molecular fingerprints~\citep{ecfp} to graph representations learned with graph convolutional and message-passing networks~\citep{gcn,mpnn}. For 3D molecular modeling, SchNet uses continuous-filter convolutions to model rotation-invariant molecular properties~\citep{schnet}. PaiNN and EGNN explicitly incorporate equivariance into graph-based geometric architectures~\citep{painn,egnn}. Recent geometric models, including ViSNet, ETNN, and GotenNet, further develop 3D molecular representation learning~\citep{visnet,etnn,gotennet}. \ours{} complements these architectures by selecting source domains and adapting their representations under structural distribution shifts, rather than introducing a new molecular backbone.

\paragraph{Molecular OOD evaluation.} Existing molecular OOD studies use dataset partitions that separate training and test molecules according to structural or other distributional criteria. MoleculeNet includes scaffold-based evaluation, in which Bemis--Murcko scaffold groups are assigned to disjoint splits~\citep{bemis_murcko,moleculenet}. GOOD provides controlled graph OOD settings for distinct forms of distribution shift~\citep{good_benchmark}, while MoleOOD studies substructure-aware invariance for learning molecular representations under OOD shifts~\citep{moleood}. DrugOOD introduces an OOD dataset curator and benchmark for drug--target binding-affinity prediction with realistic levels of annotation noise, while MOOD characterizes deployment-oriented molecular distribution shifts and examines their effects on predictive performance and uncertainty calibration~\citep{drugood,mood}. Together, these works highlight that conclusions about OOD performance depend on both how shifts are constructed and how representations are learned. Within this landscape, \bench{} evaluates molecular prediction under descriptor-based scaffold-cluster shifts by assigning clusters to source, validation, and target domains.

\paragraph{Multi-source adaptation and source selection.} Domain adaptation seeks to transfer predictive models across distribution shifts~\citep{da_theory}. Multi-source methods include adversarial alignment and moment matching~\citep{mdan,m3sda}, while Deep CORAL aligns second-order feature statistics between source and target domains~\citep{deep_coral}. In multi-source settings, individual sources can differ in transfer utility, and poorly matched sources may cause negative transfer~\citep{negative_transfer,characterizing_negative_transfer}. Graph kernels, including the Wasserstein Weisfeiler--Lehman kernel, provide structural-similarity measures~\citep{wwl_kernel}. These scores compare structures, whereas source selection must assess the effect of adapting with a set of labeled molecules. Reinforcement learning provides a general framework for combinatorial decision making~\citep{neural_combinatorial_rl}, and Group Relative Policy Optimization estimates group-relative advantages among sampled actions without training a separate value-function model~\citep{deepseekmath}. \ours{} separates candidate retrieval from utility learning. Structural similarity is used to retrieve and rank candidate sources, proxy-task adaptation performance supplies feedback for selecting source subsets, and dual-scale covariance alignment supports subsequent adaptation.

\paragraph{Transferability estimation.} Transferability measures also address the distinction between representation similarity and downstream performance. LEEP evaluates the transferability of a source classifier using predictions and labels on a target task~\citep{leep}, while LogME evaluates the evidence of target labels given features from a pretrained model~\citep{logme}. These methods assess pretrained representations without running full fine-tuning for every candidate model. Our setting differs in both the selection object and the available feedback: \ours{} selects combinations of source scaffolds, and the real target property labels are unavailable for source selection. Labeled proxy tasks provide adaptation feedback instead. Both lines of work distinguish task utility from structural similarity, but use different feedback.

\section{Method}
\label{sec:method}

\subsection{Problem Definition}
\label{subsec:problem_definition}

The prediction task is defined over 3D molecular graphs $G_i = (\mathcal{V}_i, \mathcal{E}_i, \mathbf{R}_i, \mathbf{Z}_i)$ with label $y_i \in \mathbb{R}$. The topological scaffold of each molecule is identified using Bemis--Murcko scaffold extraction, $s_i = \phi(G_i)$.

Let $\mathcal{S}$ and $\mathcal{T}$ denote the source and target scaffold sets, respectively, associated with labeled source molecules and unlabeled target molecules. We consider an extreme OOD setting in which $\mathcal{S} \cap \mathcal{T} = \emptyset$ and the two domains are separated through clustering and partitioning in an explicit physicochemical descriptor space.

Two challenges arise in this setting. First, distinct scaffolds can share local substructures, so scaffold disjointness alone does not ensure separation of local structural patterns relevant to property prediction. Second, heterogeneous source domains can differ in transfer utility, and adaptation with poorly matched sources may degrade target prediction. We therefore learn a policy to select a source subset $\mathcal{S}^* \subset \mathcal{S}$ for adaptation.

\subsection{Construction of \bench{}}
\label{subsec:scope_bench}

To construct the scaffold-cluster split based on physicochemical descriptors, \bench{} forms source, validation, and target domains through a three-step pipeline.

\textbf{Step 1: Scaffold extraction and feature construction.} Bemis--Murcko scaffolds are extracted from QM9~\citep{qm9} via RDKit~\citep{rdkit}, yielding 1,247 unique scaffolds with at least 10 samples each. Each scaffold $s_k$ is represented by a four-dimensional physicochemical descriptor:
\begin{equation}
  f_k = V_{macro} \oplus V_{element} \oplus V_{conn} \oplus V_{flex}
  \label{eq:feature_vec}
\end{equation}
where $\oplus$ denotes concatenation; $V_{macro}$ encodes global topology (atom/ring count), $V_{element}$ captures elemental polarizability, $V_{conn}$ reflects electron delocalization and rigidity, and $V_{flex}$ uses the rotatable-bond ratio as a flexibility descriptor.

\textbf{Step 2: Hierarchical clustering.} Scaffolds are first grouped into five levels by ring count and maximum ring size. Within each level, the descriptor vectors are z-score normalized and clustered with k-means++~\citep{kmeans}. Cluster quotas are allocated proportionally to level size, with a sparsity compensation coefficient $\alpha_{comp}$ for underrepresented polycyclic levels, yielding a total of $K_{total}=12$ disjoint scaffold clusters across the five level-wise normalized physicochemical descriptor spaces.

\textbf{Step 3: Asymmetric partitioning.} Six clusters form the source domain, one is used for validation, and the remaining five form the target domain. Target scaffolds with sufficient sample size are then used as target-scaffold evaluation tasks. These target scaffolds are excluded from labeled source training, while their unlabeled structures remain available for the later retrieval and adaptation steps; their property labels are not used by the source-selection policy.

The partitions are fixed by cluster assignment. The distance thresholds reported in Appendix~\ref{app:split_diagnostics} are used only for post hoc diagnostics and do not affect benchmark construction.

\subsection{Policy Optimization for Multi-source Adaptation}
\label{subsec:poma}

Under extreme structural shifts, adaptation from heterogeneous source domains may lead to negative transfer, motivating adaptive source selection. We therefore propose Policy Optimization for Multi-source Adaptation (\ours{}), which organizes molecular knowledge transfer as an integrated retrieve--compose--adapt process. Rather than relying only on fixed similarity scores, \ours{} learns a source-selection policy over combinations of candidate source domains. The policy uses structural information from the unlabeled target domain to guide source-subset selection for subsequent adaptation.

\subsubsection{Task-Specific Environment Construction}
\label{subsubsec:tsec}

In the unlabeled target adaptation setting, direct policy optimization must cope with costly proxy evaluations and a combinatorial source-subset space. To construct a task-specific surrogate feedback signal, structurally similar scaffolds from labeled source domains are selected as proxy targets using the unlabeled real-target structures.

Given $N_T$ unlabeled real target scaffolds $\mathcal{T}_{real}$ and a sufficiently large source domain pool $\mathcal{S}_{pool}$, the Morgan fingerprint~\citep{morgan} cosine similarity between a candidate scaffold $s \in \mathcal{S}_{pool}$ and a real target $t_i \in \mathcal{T}_{real}$ is computed as:
\begin{equation}
  \operatorname{Sim}(s, t_i) =
  \frac{\mathbf{v}_s \cdot \mathbf{v}_{t_i}}
       {\|\mathbf{v}_s\| \cdot \|\mathbf{v}_{t_i}\| + \epsilon}
  \label{eq:sim}
\end{equation}
where $\mathbf{v}_s$ and $\mathbf{v}_{t_i}$ denote the fingerprint feature vectors of source scaffold $s$ and target scaffold $t_i$, respectively. Here $\|\cdot\|$ is the $L_2$ norm, and $\epsilon$ is a small positive constant for numerical stability. A joint HubScore is defined by
\begin{equation}
  \operatorname{HubScore}(s) =
  \max_{t_i \in \mathcal{T}_{real}} \operatorname{Sim}(s, t_i)
  + \lambda \sum_{t_i \in \mathcal{T}_{real}}
    \mathbb{I}\!\left(\operatorname{Sim}(s, t_i) > \tau_{sim}\right)
  \label{eq:hubscore}
\end{equation}
where $\lambda>0$ is a coverage-balancing coefficient, $\mathbb{I}(\cdot)$ is the indicator function, and $\tau_{sim}$ is the similarity threshold. The top-$N_p=20$ scaffolds by HubScore form $\mathcal{T}_{proxy}$. This screening uses structural fingerprints and unlabeled target structures, not target property labels.

For each proxy target $s^*\in\mathcal{T}_{proxy}$, candidate source domains are ranked independently. Each candidate $c_j$ is a source scaffold, and $\mathcal D_{c_j}$ contains the source molecules assigned to that scaffold. The source clusters define the benchmark partition, while selection operates on scaffolds within that partition. The Weisfeiler--Lehman (WL) graph kernel~\citep{weisfeiler-lehman_kernel}, together with source-domain sample size, defines
\begin{equation}
  S_{rank}(s^*, c_j) = k_{WL}(s^*, c_j) \times
  \ln(1 + |\mathcal{D}_{c_j}|)
  \label{eq:srank}
\end{equation}
where $k_{WL}(s^*,c_j)$ is the WL similarity between proxy target $s^*$ and candidate domain $c_j$. The proxy-specific candidate pool is \[ \mathcal{C}(s^*) = \operatorname{TopM}_{c_j}\,S_{rank}(s^*,c_j), \qquad |\mathcal{C}(s^*)|=M=50. \] Each $s^*$ defines a policy-training environment with its own candidate pool and reward. Candidate pools are not merged, and rewards are not averaged across proxy tasks. One shared policy is updated sequentially across these environments.

\subsubsection{Dual-Scale Decoupled Domain Adaptation}
\label{subsubsec:msda}

To model distribution differences at complementary molecular scales, the encoder constructs two representation paths: a whole-molecule path yielding features $h_{mol}\in\mathbb{R}^d$ and a substructure path yielding BRICS-derived features $h_{sub}\in\mathbb{R}^d$ through BRICS retrosynthetic cleavage~\citep{brics}. For each path, the discrepancy between a selected source domain and the target domain is measured by the squared Frobenius norm of the difference between their empirical covariance matrices, following the Deep CORAL criterion~\citep{deep_coral}.

For the environment associated with proxy target $s^*$, let an executed action $\mathbf{a}_i$ select $K_i=\|\mathbf{a}_i\|_0$ source domains $c_1,\ldots,c_{K_i}$, and let $S_k=S_{rank}(s^*,c_k)$. For a nonempty action, the ranking scores are normalized only over its selected sources: $\gamma_k=S_k/\sum_{\ell=1}^{K_i}S_\ell$. The dual-scale alignment objective for that action is
\begin{equation}
  \mathcal{L}_{DA} =
    w_{mol} \sum_{k=1}^{K_i} \gamma_k \cdot
      \frac{\|\mathbf{C}_s^{mol,k}-\mathbf{C}_t^{mol}\|_F^2}{4d^2}
    + w_{sub} \sum_{k=1}^{K_i} \gamma_k \cdot
      \frac{\|\mathbf{C}_s^{sub,k}-\mathbf{C}_t^{sub}\|_F^2}{4d^2}
  \label{eq:da_loss}
\end{equation}
where $\mathbf{C}_s^{mol,k}$ and $\mathbf{C}_t^{mol}$ are the $d\times d$ empirical covariance matrices of whole-molecule features from selected source $k$ and the target, respectively. The corresponding substructure matrices are $\mathbf{C}_s^{sub,k}$ and $\mathbf{C}_t^{sub}$, and $\|\cdot\|_F$ denotes the Frobenius norm. All covariance matrices are estimated from the current mini-batches using the sample-covariance denominator $n-1$, with $d$ denoting the representation dimension. For the designated empty control ($K_i=0$), selected-source alignment and $\gamma$ normalization are skipped and the supervised regression path is retained. The total objective is
\begin{equation}
  \mathcal{L}_{total}=w_{reg}\,\mathcal{L}_{reg}+\mathcal{L}_{DA}.
  \label{eq:total_loss}
\end{equation}

Training proceeds in two phases. During the first $E_{warm}=10$ epochs, only $\mathcal{L}_{reg}$ is used for supervised warm-up. From epoch 11 onward, the alignment objective is activated and a backbone-specific dynamic controller updates $w_{reg}$, $w_{mol}$, and $w_{sub}$ once per epoch. The controller rules, including any applicable clipping bounds, are backbone-specific and are therefore not represented by a single shared formula.

\subsubsection{Source-domain Combinatorial Decision via GRPO}
\label{subsubsec:grpo}

For each proxy-specific environment, a state representation vector $\mathbf{x}_j\in\mathbb{R}^{d_{state}}$ with $d_{state}=258$ is constructed for candidate scaffold $c_j$:
\begin{equation}
  \mathbf{x}_j = f_{s^*} \oplus f_{c_j} \oplus
  \left[k_{WL}(s^*,c_j),
    \frac{\ln(1+|\mathcal{D}_{c_j}|)}{c_{norm}}\right]
  \label{eq:state}
\end{equation}
where $f_k$ in Eq.~\eqref{eq:feature_vec} is the four-dimensional physicochemical descriptor used for scaffold clustering, whereas $f_{s^*}$ and $f_{c_j}$ here are 128-dimensional Morgan fingerprints used by the policy state. Together with the WL similarity and normalized log sample count, these components give $d_{state}=128+128+2=258$. The shared policy network outputs a selection logit and a Bernoulli selection probability $p_j$ for each candidate.

For each policy-training group, one of the $G=33$ slots is reserved for the designated empty control $\mathbf{a}_0=\mathbf{0}$. Each remaining slot first samples a proposal $\mathbf{b}_i\in\{0,1\}^{M}$ from independent Bernoulli variables with probabilities $p_1,\ldots,p_M$. Proposals selecting at most $K_{\max}=5$ candidates are retained; proposals selecting more than five retain the entries with the largest policy logits. The resulting executed subset $\mathbf{a}_i$ has cardinality $K_i=\|\mathbf{a}_i\|_0\in[0,5]$.

For compactness, let $S=\mathbf{X}(s^*)$ denote the current proxy-specific policy state matrix.

The policy update uses the following factorized action score evaluated on the executed subset:
\begin{equation}
  \ln \pi_\theta(\mathbf{a}_i\mid S) =
  \sum_{j=1}^{M}\left[
    a_{ij}\ln p_j+(1-a_{ij})\ln(1-p_j)
  \right].
  \label{eq:logprob}
\end{equation}
Here $\pi_\theta(\mathbf{a}_i\mid S)$ denotes the factorized action score used by the policy update after the proposal has been transformed into the executed subset used by the proxy-task evaluator.

For each proxy target, all action slots in a group use the same supervised initialization and proxy-task evaluation split. A slot is considered valid only if its evaluator successfully returns a metric. The designated empty control follows the supervised path without selected-source alignment, and its returned error defines $\mathrm{MAE}_{base}$. For each valid nonempty executed subset, $\mathrm{MAE}_i$ is measured after whole-molecule and BRICS-derived substructure alignment. The raw proxy reward is \[ R_i=\mathrm{MAE}_{base}-\mathrm{MAE}_i. \] Each proxy target defines a separate policy-training task, and the same policy parameters are updated sequentially across these tasks. These rewards are surrogate feedback for source-subset selection and are not asserted to identify the optimal subset for the real target domain.

Let $V\subseteq\{0,\ldots,G-1\}$ denote the indices of action slots whose proxy evaluators successfully return a metric.

For each GRPO iteration, the $G$ slots are evaluated in parallel and the intra-group normalized advantage is
\begin{equation}
  \hat{A}_i =
  \frac{R_i-\mu_{R_V}}{\sigma_{R_V}+\epsilon_s}
  \label{eq:advantage}
\end{equation}
where $\mu_{R_V}$ and $\sigma_{R_V}$ are the mean and standard deviation of the rewards indexed by $V$.

This GRPO implementation does not train a separate critic network. To constrain policy update step sizes, a reference network $\pi_{ref}$ and a KL-based reference penalty are introduced. Letting $\Delta_i = \ln \pi_\theta(\mathbf{a}_i \mid S) - \ln \pi_{ref}(\mathbf{a}_i \mid S)$, the penalty is estimated as:
\begin{equation}
  D_{KL}\!\left(\pi_\theta \,\|\, \pi_{ref}\right) \approx
  \frac{1}{|V|} \sum_{i \in V}
  \left(\exp(\Delta_i) - \Delta_i - 1\right)
  \label{eq:kl}
\end{equation}
The GRPO optimization objective is:
\begin{equation}
  \mathcal{L}_{GRPO}(\theta) =
  -\frac{1}{|V|} \sum_{i \in V}
  \min\!\left(
    r_i(\theta)\, \hat{A}_i,
    \operatorname{clip}\!\left(r_i(\theta),
      1-\epsilon_{clip},\, 1+\epsilon_{clip}\right)
    \hat{A}_i
  \right)
  + \beta\, D_{KL}\!\left(\pi_\theta \,\|\, \pi_{ref}\right)
  \label{eq:grpo}
\end{equation}
where $r_i = \pi_\theta(\mathbf{a}_i \mid S) / \pi_{ref}(\mathbf{a}_i \mid S)$ is the score ratio used in the clipped objective, $\epsilon_{clip}$ is the clipping threshold, and $\beta > 0$ is the KL penalty coefficient. At deployment, for the current real-target scaffold $t$, candidate source domains are reranked using $S_{rank}(t,c_j)$ to construct the target-specific pool $\mathcal{C}(t)$ and state matrix $\mathbf{X}(t)$. The shared policy then deterministically selects the $K_{\mathrm{deploy}}=5$ candidates with the largest selection probabilities, and the selected source domains are used for target-specific adaptation with $\mathcal{L}_{total}$. The full staged procedure is summarized in Algorithm~\ref{alg:poma} in the appendix.

The split diagnostics are reported in Section~\ref{subsec:verification}, with the feature-space coverage definition and additional results in Appendix~\ref{app:split_diagnostics}.

\begin{figure}[t!]
  \centering
  \begin{subfigure}{0.32\textwidth}\centering
    \includegraphics[width=\linewidth]{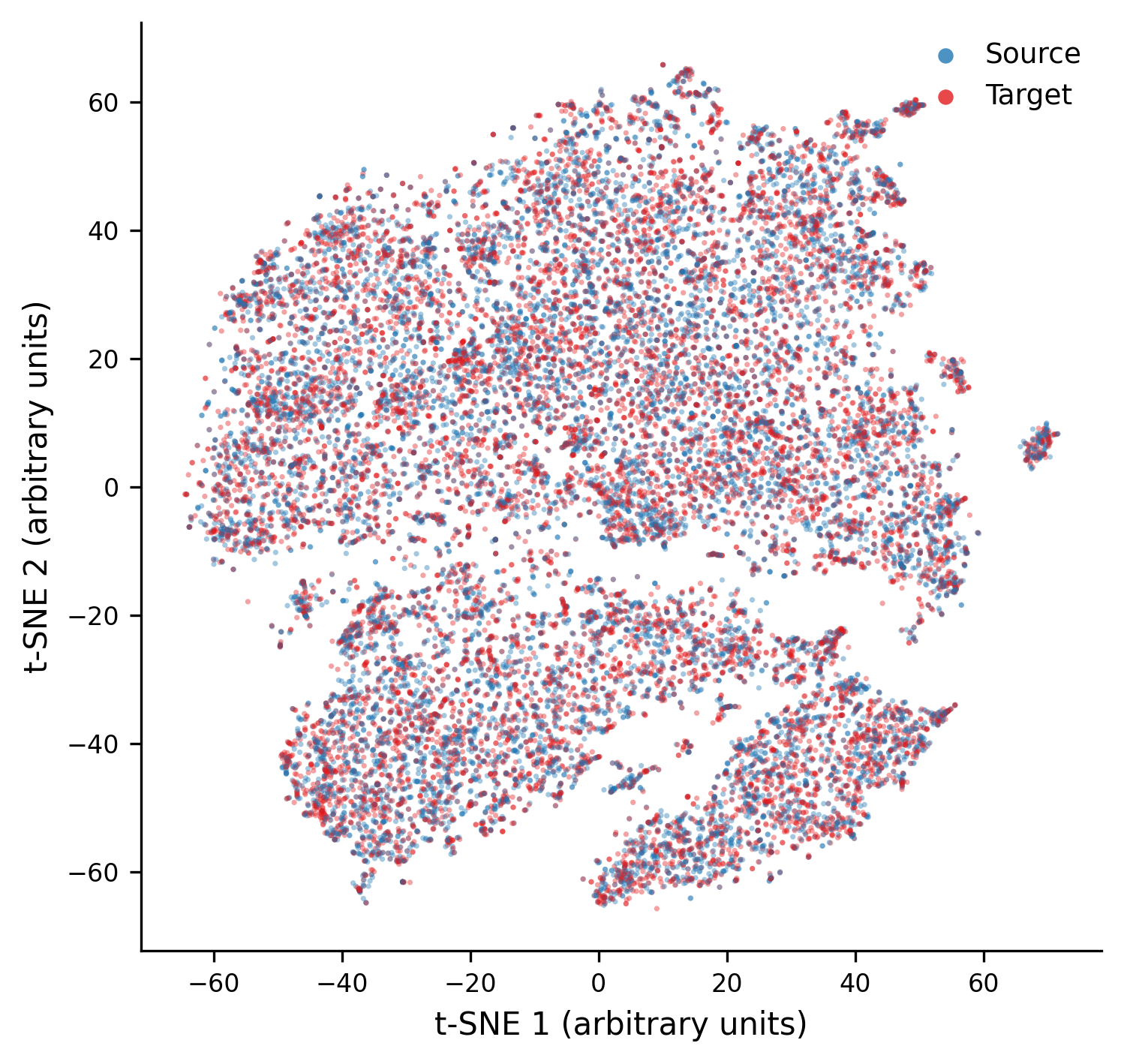}
    \caption{Random split}
  \end{subfigure}\hfill
  \begin{subfigure}{0.32\textwidth}\centering
    \includegraphics[width=\linewidth]{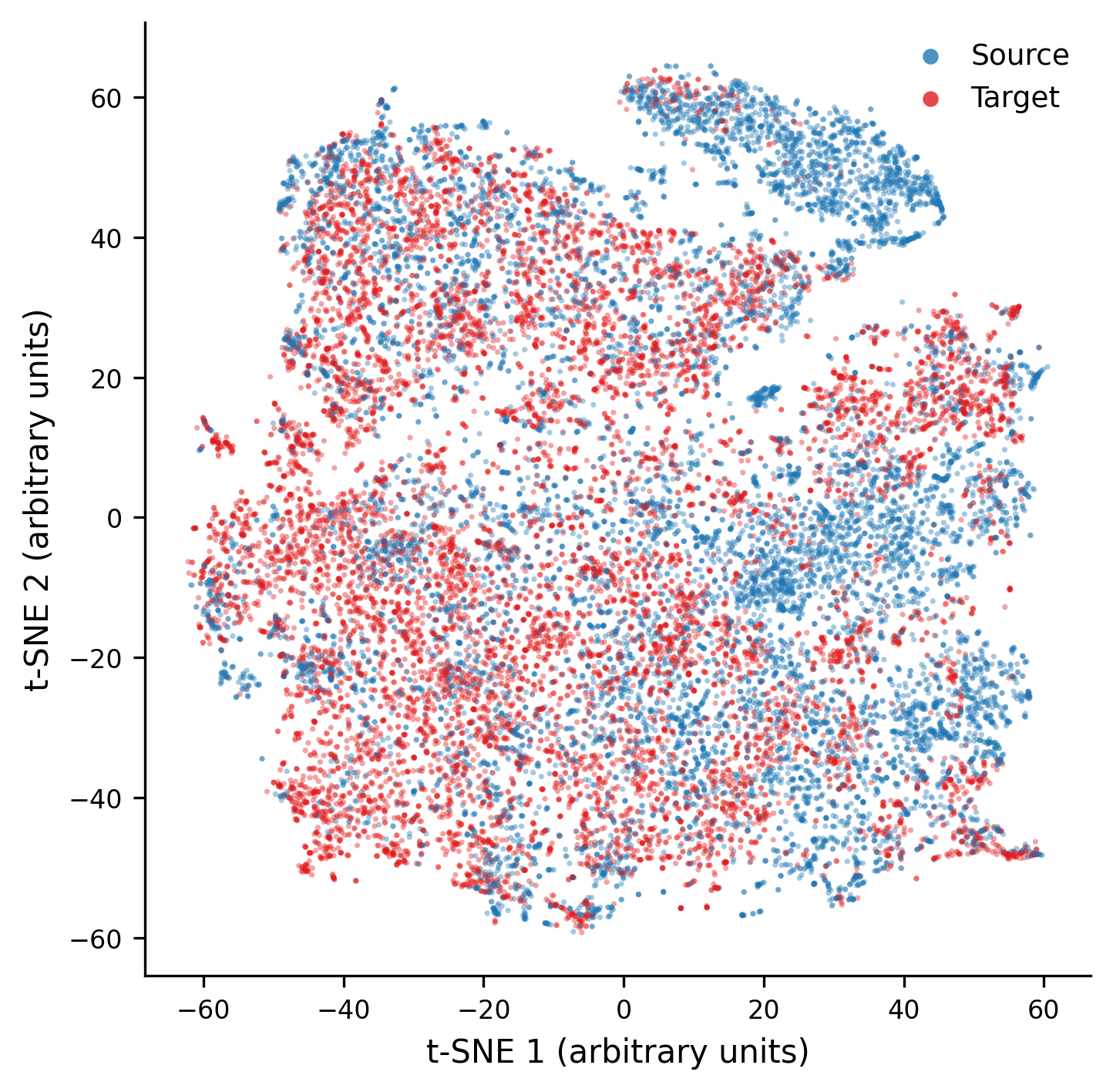}
    \caption{Standard scaffold split}
  \end{subfigure}\hfill
  \begin{subfigure}{0.32\textwidth}\centering
    \includegraphics[width=\linewidth]{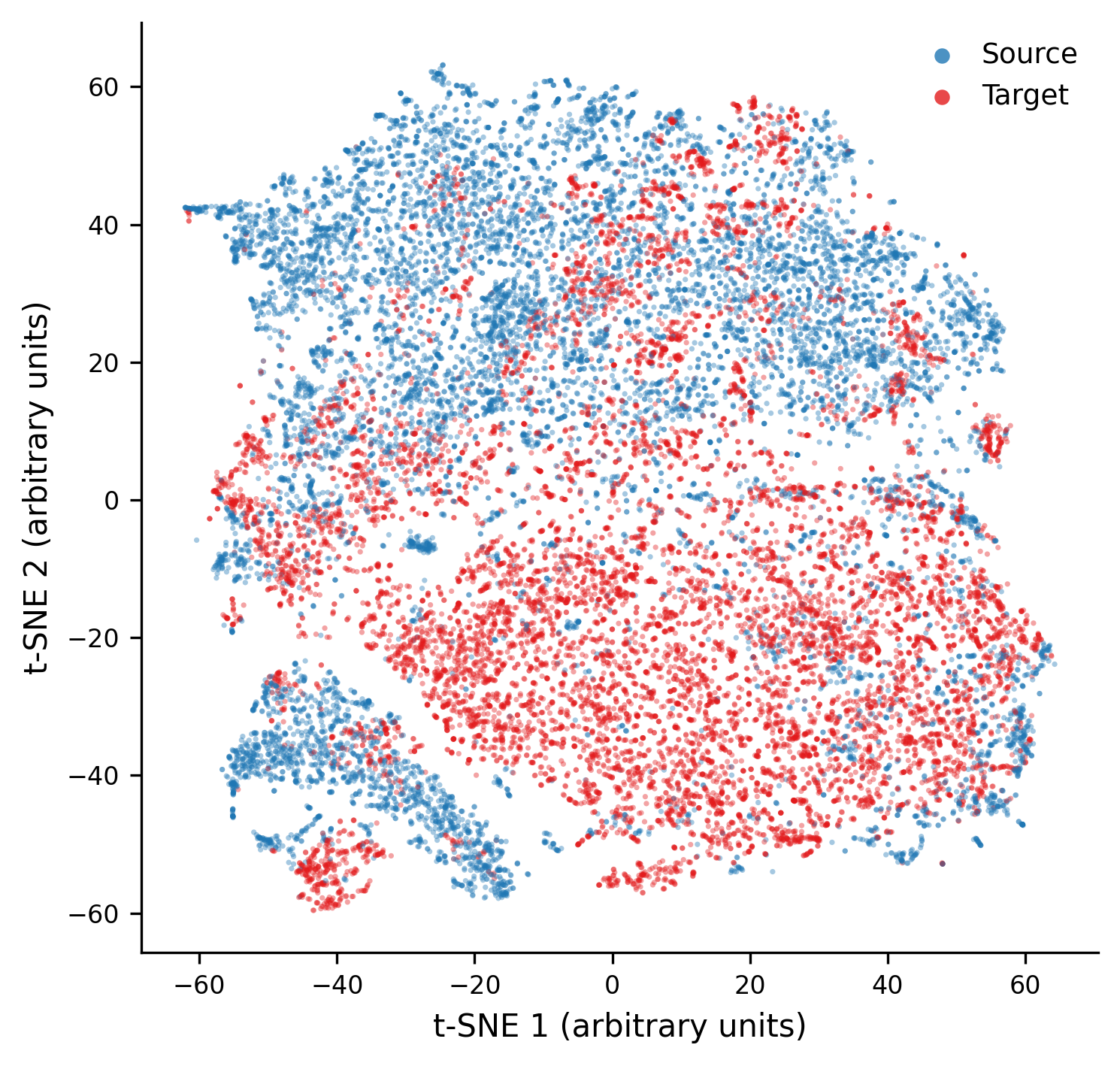}
    \caption{\bench{}}
  \end{subfigure}
  \caption{t-SNE visualizations of L2-normalized molecular features. Each protocol includes all 12,000 sampled source molecules (blue) and 12,000 sampled target molecules (red). The same feature extractor and t-SNE settings are used, but embeddings are fitted independently. Projected distances are not used to quantify separation. Appendix~\ref{app:split_diagnostics} reports diagnostics in the original feature space.}
  \label{fig:tsne_compare}
\end{figure}

\begin{table}[t!]
  \centering
  \small
\caption{Overall mean absolute error (MAE, eV) on QM9 under conventional scaffold splitting and the strict OOD split used by \bench{}. The degradation-factor row reports the strict-OOD supervised MAE divided by the corresponding scaffold-split MAE, and the final row reports the relative MAE reduction of \ours{} from the strict-OOD supervised baseline.}
  \label{tab:overall_results}
  \resizebox{\textwidth}{!}{%
  \begin{tabular}{l ccc ccc ccc}
    \toprule
    \multirow{2}{*}{}
      & \multicolumn{3}{c}{ViSNet~\citep{visnet}}
      & \multicolumn{3}{c}{ETNN~\citep{etnn}}
      & \multicolumn{3}{c}{GotenNet~\citep{gotennet}} \\
    \cmidrule(lr){2-4} \cmidrule(lr){5-7} \cmidrule(lr){8-10}
      & HOMO & LUMO & Gap
      & HOMO & LUMO & Gap
      & HOMO & LUMO & Gap \\
    \midrule
    Scaffold split
      & 0.0280 & 0.0230 & 0.0583
      & 0.0262 & 0.0238 & 0.0387
      & 0.0246 & 0.0257 & 0.0324 \\
    Strict OOD split
      & 0.1621 & 0.1829 & 0.2303
      & 0.1349 & 0.1312 & 0.1814
      & 0.1580 & 0.1902 & 0.2059 \\
    Degradation factor
      & $5.8\times$ & $8.0\times$ & $4.0\times$
      & $5.1\times$ & $5.5\times$ & $4.7\times$
      & $6.4\times$ & $7.4\times$ & $6.4\times$ \\
    \midrule
    \textbf{+}\ours{}
      & \textbf{0.1540} & \textbf{0.1673} & \textbf{0.2270}
      & \textbf{0.1233} & \textbf{0.1165} & \textbf{0.1681}
      & \textbf{0.1535} & \textbf{0.1728} & \textbf{0.2021} \\
    Improve.
      & \textcolor{red}{5.0\%} & \textcolor{red}{8.5\%} & \textcolor{red}{1.4\%}
      & \textcolor{red}{8.6\%} & \textcolor{red}{11.2\%} & \textcolor{red}{7.3\%}
      & \textcolor{red}{2.8\%} & \textcolor{red}{9.1\%} & \textcolor{red}{1.8\%} \\
    \bottomrule
  \end{tabular}}
\end{table}

\subsection{Complexity Analysis}
\label{subsec:complexity}

Let $D_{src}$ denote the number of source domains considered during retrieval, $M$ the retained candidate-pool size, and $N_{selected}$ the number of molecules in the final selected source set. Candidate construction requires scoring source domains and retaining the top $M$ candidates; its cost depends on the number and size of the source-domain graphs and the WL-kernel implementation. During policy training, each GRPO group evaluates $G$ action slots. The dominant cost is the proxy adaptation associated with each slot, which depends on the selected source data, the number of proxy-adaptation epochs and minibatches, backbone forward/backward computation, and the whole-molecule and BRICS-derived substructure alignment terms. Thus, policy training cost scales with the number of evaluated groups and action slots, not merely with candidate-pool size. After the candidate pool is fixed, the policy network computation depends on $M$; proxy and final adaptation costs continue to depend on the data processed for the selected sources. At deployment, deterministic top-$K_{\mathrm{deploy}}$ selection operates over the $M$ candidates, followed by target-specific adaptation on the selected source data. Per-query prediction then incurs the cost of the chosen molecular backbone; retrieval, selection, and adaptation are additional per-target setup costs.

\section{Experiments}
\label{sec:experiments}

\subsection{Experimental Setup}
\label{subsec:setup}

\bench{} is built on QM9~\citep{qm9} following the protocol of Section~\ref{subsec:scope_bench}. Three quantum chemical properties are evaluated: the highest occupied molecular orbital energy (HOMO), the lowest unoccupied molecular orbital energy (LUMO), and HOMO--LUMO gap (Gap), all in eV.
All results are obtained with a fixed random seed of 42 for reproducibility. Two protocols are compared: a supervised baseline fine-tuned on the source domain for 200 epochs without adaptation, and \ours{} with offline policy optimization followed by target-specific adaptation.

\subsection{Verification of SCOPE-Bench}
\label{subsec:verification}

\bench{} partitions 133,885 QM9 molecules into 12 disjoint scaffold clusters: 6 form the source domain (94,562 samples), 1 the validation set (18,326 samples), and 5 the target domain (19,894 samples), with 15 target-scaffold evaluation tasks based on scaffolds with $N \geq 200$. The t-SNE panels retain equal numbers of sampled source and target molecules across protocols (Figure~\ref{fig:tsne_compare}). Pairwise Tanimoto similarity~\citep{tanimoto} heatmaps provide a complementary qualitative view of cross-domain structural similarity (Figure~\ref{fig:heatmap_compare} in Appendix~\ref{app:split_diagnostics}).

For the available diagnostic feature caches, we additionally measure nearest-source coverage in a common ViSNet representation before dimensionality reduction. At the respective 95th-percentile source-calibrated radii, the target coverage rates are 94.63\%, 86.00\%, and 69.36\% for caches labeled random, conventional scaffold, and \bench{}, respectively. Their ordering is unchanged at the 90th and 99th calibration percentiles. These cache-level measurements characterize proximity in the specified representation rather than independently validating the full benchmark partitions. Appendix~\ref{app:split_diagnostics} gives the distance definition, actual radii, and the limits of the available cache provenance. The qualitative figures and coverage measurements do not establish complete chemical or substructure isolation.

Prediction error provides a separate comparison. In the top section of Table~\ref{tab:overall_results}, the MAE under \bench{} exceeds the corresponding conventional scaffold-split MAE across all nine evaluated backbone--property combinations, with a mean degradation factor of $5.9\times$. This error comparison and the representation diagnostics describe different aspects of the evaluated shifts.

\begin{table}[t!]
\centering
\small
\caption{ViSNet--HOMO ablations on \bench{}. The upper block compares source selection policies with full dual-scale adaptation. The lower block removes one alignment module while keeping the GRPO-learned policy fixed. Positive and negative $\Delta$ values indicate improvement and deterioration relative to the supervised baseline, respectively. Symbols $\checkmark$ and $\times$ denote enabled and disabled modules.}
\label{tab:ablation}
\begin{tabular}{lcccc}
\toprule
Policy & Mol-CORAL (M) & Sub-CORAL (S) & MAE & $\Delta$ \\
\midrule
Only ViSNet & - & - & 0.1621 & - \\
\midrule
Random selection         & $\checkmark$ & $\checkmark$ & 0.1708 & $-5.3\%$ \\
Shallow feature matching & $\checkmark$ & $\checkmark$ & 0.1663 & $-2.5\%$ \\
Deep feature matching    & $\checkmark$ & $\checkmark$ & 0.1669 & $-2.9\%$ \\
Physical descriptor      & $\checkmark$ & $\checkmark$ & 0.1641 & $-1.2\%$ \\
Graph kernel similarity  & $\checkmark$ & $\checkmark$ & 0.1616 & $+0.3\%$ \\
Mixed strategy           & $\checkmark$ & $\checkmark$ & 0.1576 & $+2.8\%$ \\
\midrule
\ours{} w/o M   & $\times$ & $\checkmark$ & 0.1689 & $-4.2\%$ \\
\ours{} w/o S   & $\checkmark$ & $\times$ & 0.1672 & $-3.1\%$ \\
\midrule
\ours{} (Full)           & $\checkmark$ & $\checkmark$ & 0.1540 & $\mathbf{+5.0\%}$ \\
\bottomrule
\end{tabular}
\end{table}

\subsection{Overall Results}
\label{subsec:main_results}

We evaluate \ours{} on the 15 target-scaffold tasks of \bench{} on QM9 using three backbone architectures, with the supervised model trained on all available source data without adaptation as the baseline. Table~\ref{tab:overall_results} reports the main results, and Appendix~\ref{app:scaffold_analysis} provides per-scaffold results for each backbone--property combination.

As shown in the bottom section of Table~\ref{tab:overall_results}, \ours{} achieves lower reported MAE for all nine evaluated backbone--property combinations. For HOMO, the main-table entries are 5.0\%, 8.6\%, and 2.8\% for ViSNet, ETNN, and GotenNet, respectively. For LUMO, the corresponding entries are 8.5\%, 11.2\%, and 9.1\%. For the HOMO--LUMO gap, the ViSNet and GotenNet entries are 1.4\% and 1.8\%, respectively; the ETNN--Gap entry in the main table reports a 7.3\% relative MAE reduction.
Each relative reduction compares \ours{} with the corresponding strict-OOD supervised entry, rather than with the conventional scaffold-split model. The degradation factors and adaptation gains therefore have different reference points and answer different evaluation questions. The reported results use inherited backbone hyperparameters without backbone-specific tuning.

\subsection{Ablation Study}
\label{subsec:ablation}

Learned source selection and dual-scale adaptation are the two core components of \ours{}.

\textit{\textbf{Q1: How does learned source selection compare with the evaluated heuristics?}} We compare six non-RL heuristics on ViSNet while keeping the full adaptation module fixed. Random selection provides a baseline without learned source selection. Shallow feature matching uses cosine similarity of graph embeddings after 40 epochs, whereas deep feature matching uses embeddings after 200 epochs. The physical-descriptor strategy ranks sources by atomic composition and orbital energy penalties. Graph-kernel similarity uses the WL kernel, and the mixed strategy linearly combines WL and physical scores. In Table~\ref{tab:ablation}, deep feature matching gives an MAE of 0.1669, compared with 0.1621 for the supervised baseline. \ours{} gives a lower reported mean MAE than the strongest evaluated heuristic, with values of 0.1540 and 0.1576, respectively. The comparison isolates the evaluated selection strategies under a common adaptation design.

\textit{\textbf{Q2: What is the effect of removing each alignment scale?}}
As shown in the lower block of Table~\ref{tab:ablation}, removing Mol-CORAL or Sub-CORAL increases aggregate MAE from 0.1540 for the full model to 0.1689 and 0.1672, respectively; both ablated values are above the supervised baseline of 0.1621. These results indicate that both alignment components contribute in the evaluated ViSNet--HOMO setting. The per-scaffold comparisons show that retaining both alignment scales does not yield the lowest error for every target. For Scaffold 4, removing Sub-CORAL gives an MAE of 0.1473, compared with 0.1576 for the full model (Appendix~\ref{app:detailed_ablation}).

\begin{figure}[t]
  \centering
  \small
  \begin{subfigure}{0.48\textwidth}
    \centering
    \includegraphics[width=\linewidth]{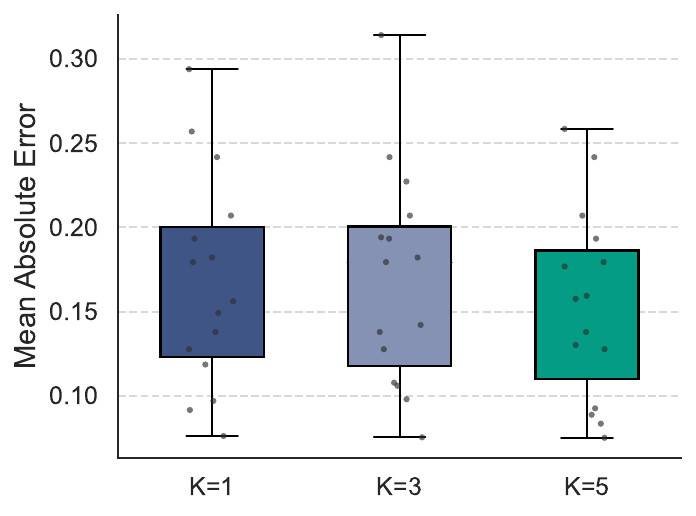}
    \caption{MAE versus subset size $K$ at $M=50$.}
    \label{fig:hp_k}
  \end{subfigure}\hfill
  \begin{subfigure}{0.48\textwidth}
    \centering
    \includegraphics[width=\linewidth]{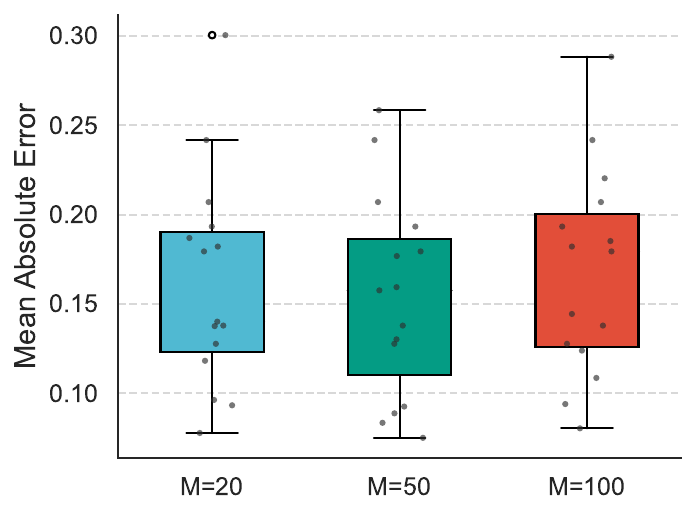}
    \caption{MAE versus pool size $M$ at $K=5$.}
    \label{fig:hp_m}
  \end{subfigure}
  \caption{Hyperparameter sensitivity on the HOMO task for ViSNet using \bench{}. Box plots summarize MAE across the 15 target scaffolds.}
  \label{fig:hp_analysis}
\end{figure}

\subsection{Hyperparameter Analysis}
\label{subsec:sensitivity}

The sensitivity of \ours{} to candidate pool size $M$ and subset size $K$ is evaluated on the HOMO task using the ViSNet architecture. The corresponding box plots are shown in Figures~\ref{fig:hp_k} and~\ref{fig:hp_m}. The default $K=5$ setting achieves the lowest MAE among the compared values in this analysis. With $K=5$ fixed, the default $M=50$ setting provides the lowest MAE among the evaluated candidate-pool sizes and is used in the main experiments.

The box plots summarize variation across target scaffolds, with the individual values reported in Appendix~\ref{app:hyperparameter_analysis}. They describe heterogeneity among the evaluated tasks rather than uncertainty across repeated training runs. The default 40 GRPO steps provide the reported performance under the evaluated setting, while increasing the number of steps yields only limited additional improvement with higher computational cost.

\subsection{Discussion and scope}
\label{subsec:discussion}

Source selection improves the reported aggregate errors but does not remove the performance gap associated with the stricter evaluation. The per-scaffold tables show that an aggregate benefit can coexist with deterioration on individual targets. The experiments therefore support adaptive source selection within the evaluated QM9 setting, rather than a claim that every target benefits from adaptation.

The study uses one fixed training seed and one scaffold-cluster partition. Variation across scaffolds describes task heterogeneity, not variability across training runs or alternative partitions. Proxy rewards measure adaptation utility on source-derived tasks rather than optimality on the real target.

\section{Conclusion}
\label{sec:conclusion}

In this paper, we introduced \bench{} for evaluating molecular prediction under descriptor-based scaffold-cluster shifts and developed \ours{}, a policy-driven pipeline combining source selection with dual-scale covariance alignment. The policy uses proxy-task adaptation feedback to select source subsets for unlabeled target adaptation. On \bench{}, the mean degradation factor relative to conventional scaffold splits is $5.9\times$; relative to the supervised fine-tuning baseline, \ours{} reduces MAE by an average of 6.2\% over the nine evaluated backbone--property combinations. A current limitation is the computational cost of the GRPO training loop, which requires executing multiple proxy domain adaptation procedures per policy iteration. In future work, surrogate reward models will be explored to amortize this cost, and the framework will be extended to zero-proxy target scenarios.

\clearpage
\subsection*{AI use statement}

Generative AI tools were used during manuscript preparation for language editing, formatting, consistency checking, proofreading, and structured review of methodological and experimental descriptions. All research decisions, implementations, experiments, results, and conclusions were developed and verified by the authors. The authors reviewed all AI-assisted outputs and take responsibility for the final content.

\subsection*{Reproducibility statement}

The benchmark construction and evaluation protocol are described in the main paper. Algorithmic and implementation details, hyperparameter settings, ablation results, and complete per-scaffold results are provided in the appendix. Code is available at the repository linked in the abstract.

{ \small \setlength{\bibsep}{2pt plus 0.3pt minus 0.3pt}
\bibliographystyle{iclr2027_conference} \bibliography{refs} }

\clearpage

\appendix

\section{Training Algorithm and Implementation Details}
\label{app:algorithm}

All experiments are conducted on NVIDIA RTX A6000 (48\,GB) GPUs. The computational overhead of \ours{} can be decoupled into two phases: (1) a one-time offline GRPO policy optimization, requiring approximately 72 GPU hours to converge; and (2) an online target adaptation phase, which leverages the pre-trained policy for source selection and dual-scale alignment. Under the default configuration ($M=50$, $K=5$, 40 GRPO steps), the online phase requires approximately 4.5 GPU hours per target scaffold.

\begin{algorithm}[H]
\caption{\ours{}: Shared-policy Proxy Training and Target-specific Adaptation}
\label{alg:poma}
\begin{algorithmic}[1]
\Require Source pool $\mathcal{S}_{pool}$, unlabeled target scaffolds
  $\mathcal{T}_{real}$, backbone $f_\theta$, policy $\pi_\theta$
\Ensure Model adapted without target property labels
\State Compute $\operatorname{HubScore}(s)$ for each
  $s\in\mathcal{S}_{pool}$ using Equation~\eqref{eq:hubscore}
\State Select the top-$N_p=20$ scaffolds as $\mathcal{T}_{proxy}$
\State Use one shared policy $\pi_\theta$ across all proxy tasks
\For{each proxy target $s^*\in\mathcal{T}_{proxy}$}
  \State Rank candidate domains using $S_{rank}(s^*,c_j)$
  \State Construct the proxy-specific pool
    $\mathcal{C}(s^*)=\operatorname{TopM}_{c_j}S_{rank}(s^*,c_j)$,
    with $M=50$
  \State Construct state matrix $\mathbf{X}(s^*)$
  \For{each GRPO iteration for this proxy task}
    \State Reserve one of the $G=33$ slots as
      $\mathbf{a}_0=\mathbf{0}$, the designated empty control
    \State Sample $G-1$ independent Bernoulli proposals
      $\{\mathbf{b}_1,\ldots,\mathbf{b}_{G-1}\}$
    \State Apply the $K_{\max}=5$ cardinality transform to obtain executed
      subsets $\mathbf{a}_1,\ldots,\mathbf{a}_{G-1}$
    \State Set $K_i=\|\mathbf{a}_i\|_0$ for each executed proposal
    \State Evaluate all $G$ slots on the common proxy initialization and
      evaluation split; a slot is valid only if its evaluator returns a metric
    \State If the designated control is valid, include it in the runtime-valid
      set used for group statistics and the implementation-level update; it is
      a fixed control slot rather than a Bernoulli proposal
    \State Define $\mathrm{MAE}_{base}$ from the designated empty control
    \State For each valid nonempty slot, compute
      $R_i=\mathrm{MAE}_{base}-\mathrm{MAE}_i$ after dual-scale adaptation
    \State Compute group-relative advantages and update the shared $\pi_\theta$ using
      Equations~\eqref{eq:kl}--\eqref{eq:grpo}
  \EndFor
\EndFor
\State At deployment, for the current real-target scaffold $t$, rank candidate
  domains using $S_{rank}(t,c_j)$
\State Construct $\mathcal{C}(t)=\operatorname{TopM}_{c_j}S_{rank}(t,c_j)$
  and the target-specific state matrix $\mathbf{X}(t)$
\State Apply the shared policy to $\mathbf{X}(t)$ and select the deterministic
  top-$K_{\mathrm{deploy}}=5$ candidates by selection probability
\State Adapt $f_\theta$ on the selected source domains with
  $\mathcal{L}_{total}$
\State \Return the adapted model and its target predictions
\end{algorithmic}
\end{algorithm}

\section{Detailed Performance Analysis across Individual Scaffolds}
\label{app:scaffold_analysis}

Table~\ref{tab:detailed_scaffold_results} reports per-scaffold MAE for the supervised baseline and \ours{} across three backbones and three properties. Relative performance varies across scaffolds and backbone--property combinations: \ours{} has lower MAE in many entries, while the baseline MAE is lower or equal in others. The table therefore provides a granular view of performance heterogeneity across the 15 target scaffolds.

\begin{table}[!htb]
  \centering
  \small
\caption{Detailed mean absolute error (MAE) on each of the 15 individual target scaffolds of \bench{}. For each scaffold and property, the better performance between the supervised baseline and \ours{} is highlighted in bold.}
  \label{tab:detailed_scaffold_results}
  \setlength{\tabcolsep}{3pt}
  \renewcommand{\arraystretch}{0.95}
  \resizebox{\textwidth}{!}{%
  \begin{tabular}{ll ccc ccc ccc}
    \toprule
    \multirow{2}{*}{Target} & \multirow{2}{*}{Method}
      & \multicolumn{3}{c}{ViSNet~\citep{visnet}}
      & \multicolumn{3}{c}{ETNN~\citep{etnn}}
      & \multicolumn{3}{c}{GotenNet~\citep{gotennet}} \\
    \cmidrule(lr){3-5} \cmidrule(lr){6-8} \cmidrule(lr){9-11}
      & & HOMO$\downarrow$ & LUMO$\downarrow$ & Gap$\downarrow$
      & HOMO$\downarrow$ & LUMO$\downarrow$ & Gap$\downarrow$
      & HOMO$\downarrow$ & LUMO$\downarrow$ & Gap$\downarrow$ \\
    \midrule

    \multirow{2}{*}{Scaffold 1}
      & supervised baseline & 0.1469 & 0.1165 & 0.1941 & 0.1138 & 0.0944 & 0.1555 & 0.1139 & 0.0897 & \textbf{0.1446} \\
      & \ours{}  & \textbf{0.1302} & \textbf{0.1081} & \textbf{0.1625} & \textbf{0.1077} & \textbf{0.0912} & \textbf{0.1455} & \textbf{0.1044} & \textbf{0.0833} & 0.1640 \\
    \midrule

    \multirow{2}{*}{Scaffold 2}
      & supervised baseline & 0.1019 & \textbf{0.0987} & \textbf{0.1206} & \textbf{0.0663} & \textbf{0.0895} & \textbf{0.1311} & \textbf{0.0748} & \textbf{0.0909} & \textbf{0.1154} \\
      & \ours{}  & \textbf{0.0835} & 0.1008 & 0.1627 & 0.0706 & 0.1243 & 0.1417 & 0.0774 & 0.0928 & 0.1242 \\
    \midrule

    \multirow{2}{*}{Scaffold 3}
      & supervised baseline & 0.1277 & 0.1177 & 0.1926 & \textbf{0.1481} & 0.1483 & \textbf{0.1601} & 0.1366 & 0.1819 & 0.1524 \\
      & \ours{}  & 0.1277 & 0.1177 & 0.1926 & 0.1515 & \textbf{0.1301} & 0.1886 & 0.1366 & \textbf{0.1503} & 0.1524 \\
    \midrule

    \multirow{2}{*}{Scaffold 4}
      & supervised baseline & 0.2032 & 0.1890 & \textbf{0.1894} & 0.1035 & \textbf{0.1626} & 0.1875 & \textbf{0.1321} & \textbf{0.1775} & 0.2245 \\
      & \ours{}  & \textbf{0.1576} & \textbf{0.1819} & 0.1945 & \textbf{0.0959} & 0.1635 & \textbf{0.1529} & 0.1406 & 0.2160 & \textbf{0.1792} \\
    \midrule

    \multirow{2}{*}{Scaffold 5}
      & supervised baseline & 0.1794 & 0.1923 & 0.1868 & 0.2205 & 0.1639 & \textbf{0.1765} & 0.2019 & 0.3214 & 0.1966 \\
      & \ours{}  & 0.1794 & 0.1923 & 0.1868 & \textbf{0.1390} & \textbf{0.0922} & 0.2103 & 0.2019 & \textbf{0.2948} & 0.1966 \\
    \midrule

    \multirow{2}{*}{Scaffold 6}
      & supervised baseline & 0.1121 & 0.1629 & 0.1804 & \textbf{0.1000} & 0.0859 & 0.1185 & 0.1077 & 0.0849 & \textbf{0.1057} \\
      & \ours{}  & \textbf{0.0926} & \textbf{0.1333} & \textbf{0.1314} & 0.1030 & \textbf{0.0792} & \textbf{0.1134} & \textbf{0.1011} & \textbf{0.0847} & 0.1274 \\
    \midrule

    \multirow{2}{*}{Scaffold 7}
      & supervised baseline & 0.2697 & 0.4440 & 0.7133 & 0.1704 & 0.2429 & 0.4447 & 0.2333 & 0.3014 & 0.5344 \\
      & \ours{}  & \textbf{0.2584} & \textbf{0.3014} & \textbf{0.7077} & \textbf{0.1436} & \textbf{0.2242} & \textbf{0.1710} & \textbf{0.2173} & 0.3014 & \textbf{0.4900} \\
    \midrule

    \multirow{2}{*}{Scaffold 8}
      & supervised baseline & 0.0901 & \textbf{0.0871} & \textbf{0.1026} & \textbf{0.0775} & 0.0641 & \textbf{0.0861} & \textbf{0.0825} & 0.0669 & 0.0907 \\
      & \ours{}  & \textbf{0.0888} & 0.0923 & 0.1050 & 0.0817 & \textbf{0.0502} & 0.0891 & 0.0841 & \textbf{0.0650} & \textbf{0.0899} \\
    \midrule

    \multirow{2}{*}{Scaffold 9}
      & supervised baseline & 0.2417 & 0.1897 & 0.2893 & 0.2486 & 0.1332 & \textbf{0.2047} & 0.3014 & 0.2630 & 0.2441 \\
      & \ours{}  & 0.2417 & 0.1897 & 0.2893 & \textbf{0.2442} & \textbf{0.1312} & 0.2288 & 0.3014 & \textbf{0.1920} & 0.2441 \\
    \midrule

    \multirow{2}{*}{Scaffold 10}
      & supervised baseline & 0.1379 & 0.1978 & 0.2100 & \textbf{0.1207} & 0.1377 & 0.1674 & 0.1400 & 0.1927 & 0.1331 \\
      & \ours{}  & 0.1379 & 0.1978 & 0.2100 & 0.1254 & \textbf{0.1038} & \textbf{0.1159} & 0.1400 & \textbf{0.1436} & 0.1331 \\
    \midrule

    \multirow{2}{*}{Scaffold 11}
      & supervised baseline & 0.1821 & 0.2451 & 0.2137 & 0.1607 & 0.2050 & \textbf{0.1496} & 0.1960 & 0.2615 & 0.2930 \\
      & \ours{}  & \textbf{0.1768} & 0.2451 & 0.2137 & \textbf{0.1490} & \textbf{0.1253} & 0.2421 & 0.1960 & \textbf{0.2113} & 0.2930 \\
    \midrule

    \multirow{2}{*}{Scaffold 12}
      & supervised baseline & 0.1933 & 0.1821 & 0.2650 & \textbf{0.1328} & \textbf{0.0981} & \textbf{0.1907} & 0.2211 & 0.1474 & 0.1989 \\
      & \ours{}  & 0.1933 & 0.1821 & 0.2650 & 0.1415 & 0.0986 & 0.1992 & 0.2211 & \textbf{0.1276} & 0.1989 \\
    \midrule

    \multirow{2}{*}{Scaffold 13}
      & supervised baseline & 0.2070 & 0.1576 & 0.2254 & 0.1367 & 0.1287 & \textbf{0.1739} & 0.2184 & 0.1414 & 0.1869 \\
      & \ours{}  & 0.2070 & 0.1576 & 0.2254 & \textbf{0.1247} & \textbf{0.1175} & 0.1886 & 0.2184 & \textbf{0.1179} & 0.1869 \\
    \midrule

    \multirow{2}{*}{Scaffold 14}
      & supervised baseline & \textbf{0.0574} & 0.0969 & \textbf{0.0921} & 0.0722 & 0.0862 & \textbf{0.1059} & 0.0661 & \textbf{0.0854} & 0.0891 \\
      & \ours{}  & 0.0751 & \textbf{0.0909} & 0.1008 & \textbf{0.0644} & \textbf{0.0845} & 0.1163 & \textbf{0.0642} & 0.0981 & \textbf{0.0826} \\
    \midrule

    \multirow{2}{*}{Scaffold 15}
      & supervised baseline & 0.1818 & 0.2653 & 0.2799 & 0.1515 & \textbf{0.1272} & 0.2235 & 0.1445 & 0.4472 & 0.3784 \\
      & \ours{}  & \textbf{0.1594} & \textbf{0.2188} & \textbf{0.2581} & \textbf{0.1076} & 0.1317 & \textbf{0.2176} & \textbf{0.0985} & \textbf{0.4138} & \textbf{0.3689} \\
    \bottomrule
  \end{tabular}}
\end{table}

\FloatBarrier
\section{Feature-space Diagnostics of the Split Protocols}
\label{app:split_diagnostics}

This analysis is a post hoc diagnostic of fixed splits, not a criterion used to construct them. For each protocol-specific diagnostic cache, we sample 12,000 source and 12,000 target molecules without replacement. The source sample is divided into disjoint reference and calibration sets of 6,000 molecules each. We use the same fixed ViSNet checkpoint and its 128-dimensional molecular feature output for all three caches, and normalize each feature vector to unit $L_2$ norm. The diagnostic caches use row-level identifiers, and their mappings to the formal split counts and molecule identities have not been independently established. The diagnostic therefore describes the specified cache samples and does not by itself verify target-label non-exposure or encoder-training provenance.

\begin{figure}[!htb]
  \centering
  \includegraphics[width=0.78\linewidth]{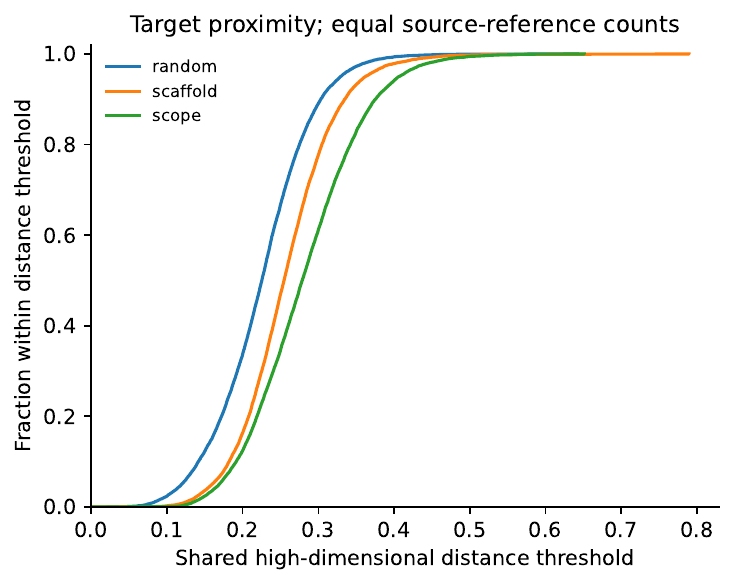}
  \caption{Empirical cumulative distributions of nearest-source distances in the shared $L_2$-normalized ViSNet representation. Each diagnostic cache uses 12,000 target queries and 6,000 source references. A curve gives the fraction of queries within each distance threshold. The horizontal axis is shared, unlike the separately calibrated radii in Table~\ref{tab:source_coverage}. These distances describe representation proximity, not chemical substructure overlap.}
  \label{fig:source_distance_ecdf}
\end{figure}

Let $p$ index a protocol, and let $\mathcal{R}_p$, $\mathcal{A}_p$, and $\mathcal{T}^{\mathrm{diag}}_p$ denote its source reference, source calibration, and target diagnostic samples. With $\widetilde h(x)=h(x)/\|h(x)\|_2$, define
\[
 d_p(x)=\min_{s\in\mathcal{R}_p}\|\widetilde h(x)-\widetilde h(s)\|_2,
 \qquad
 \rho_{p,q}=\operatorname{Quantile}_q\{d_p(s):s\in\mathcal{A}_p\}.
\]
The corresponding target coverage is
\[
 \widehat C_p(q)=\frac{1}{|\mathcal{T}^{\mathrm{diag}}_p|}
 \sum_{t\in\mathcal{T}^{\mathrm{diag}}_p}
 \mathbb{I}\!\left[d_p(t)\leq\rho_{p,q}\right].
\]
We report $q\in\{0.90,0.95,0.99\}$ using linear interpolation to compute quantiles. These are calibration quantiles, not confidence levels. The resulting radii are used only for post hoc diagnostics and do not determine the split. Figure~\ref{fig:source_distance_ecdf} shows the empirical cumulative distributions of $d_p(t)$ on a common absolute distance axis.

\begin{table}[!htb]
\centering
\small
\caption{Post hoc target coverage in the L2-normalized 128-dimensional ViSNet feature space for the diagnostic caches. Each cache uses 6,000 source reference molecules, 6,000 source calibration molecules, and 12,000 target queries. The radius $\rho_{p,q}$ is calibrated separately for each cache at source-distance quantile $q$; $q$ is not a confidence level.}
\label{tab:source_coverage}
\setlength{\tabcolsep}{4pt}
\renewcommand{\arraystretch}{0.95}
\begin{tabular}{lcccc}
\toprule
Protocol & $q$ & $\rho_{p,q}$ & Covered targets & Coverage \\
\midrule
Random & 0.90 & 0.30052 & 10,706 / 12,000 & 89.22\% \\
Scaffold & 0.90 & 0.29849 & 9,236 / 12,000 & 76.97\% \\
\bench{} & 0.90 & 0.29129 & 6,816 / 12,000 & 56.80\% \\
\midrule
Random & 0.95 & 0.32693 & 11,356 / 12,000 & 94.63\% \\
Scaffold & 0.95 & 0.32076 & 10,320 / 12,000 & 86.00\% \\
\bench{} & 0.95 & 0.31643 & 8,323 / 12,000 & 69.36\% \\
\midrule
Random & 0.99 & 0.38111 & 11,867 / 12,000 & 98.89\% \\
Scaffold & 0.99 & 0.37505 & 11,550 / 12,000 & 96.25\% \\
\bench{} & 0.99 & 0.38116 & 10,946 / 12,000 & 91.22\% \\
\bottomrule
\end{tabular}
\end{table}

At $q=0.95$, the protocol-specific diagnostic caches give radii of 0.32693, 0.32076, and 0.31643 for random, conventional scaffold, and \bench{}, respectively. The corresponding coverage rates are 94.63\%, 86.00\%, and 69.36\%; the median target-to-reference distances are 0.22535, 0.25454, and 0.27861. The same coverage ordering holds at $q=0.90$ and $q=0.99$ (Table~\ref{tab:source_coverage}). These are descriptive measurements for the specified samples, representation, and distance metric. They do not establish complete chemical or substructure isolation, invariance to alternative descriptor or clustering choices, or statistical significance of prediction results.

The t-SNE plots~\citep{tsne} in Figure~\ref{fig:tsne_compare} are fitted independently to the concatenated source and target samples for each cache. We use perplexity 55, PCA initialization, learning rate 200, early exaggeration 12, Barnes--Hut angle 0.5, a maximum of 1,000 iterations, and random state 42. All sampled points are plotted with equal marker size and opacity; no local-purity or density mask is applied. Because the embeddings are fitted independently, their coordinates are not interpreted as a common distance scale.

\begin{figure}[!htb]
  \centering
  \begin{subfigure}{0.32\textwidth}\centering
    \includegraphics[width=\linewidth]{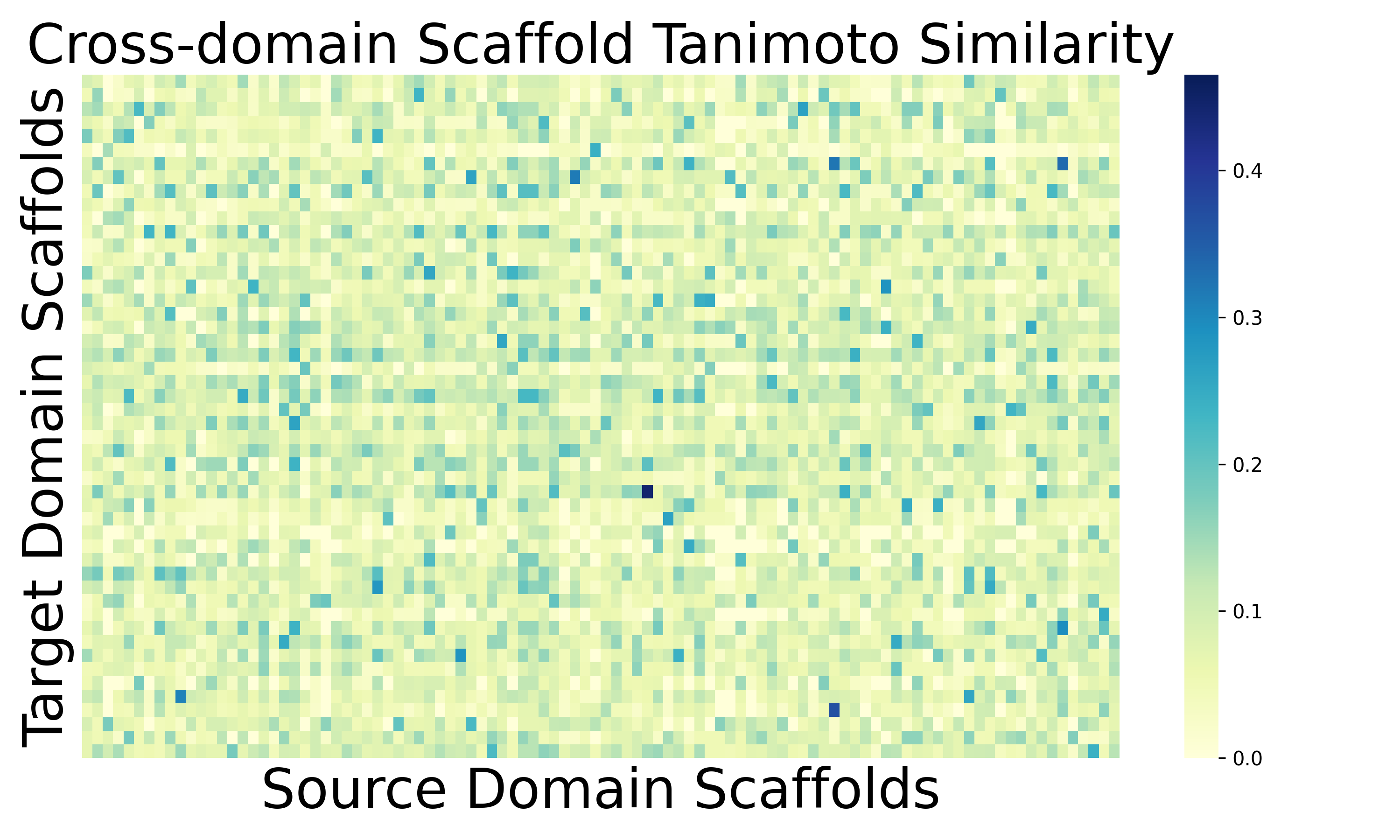}
    \caption{Random split}
  \end{subfigure}\hfill
  \begin{subfigure}{0.32\textwidth}\centering
    \includegraphics[width=\linewidth]{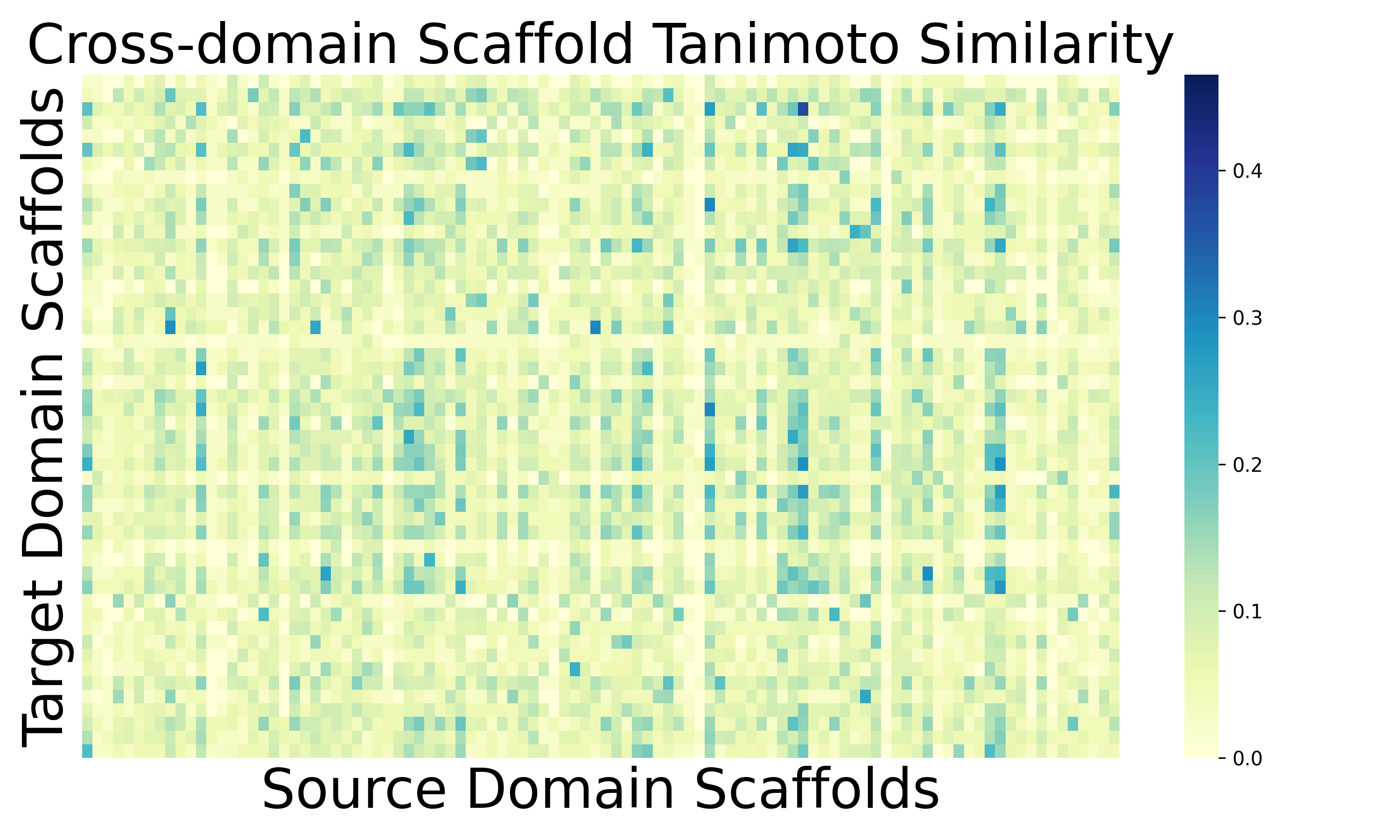}
    \caption{Standard scaffold split}
  \end{subfigure}\hfill
  \begin{subfigure}{0.32\textwidth}\centering
    \includegraphics[width=\linewidth]{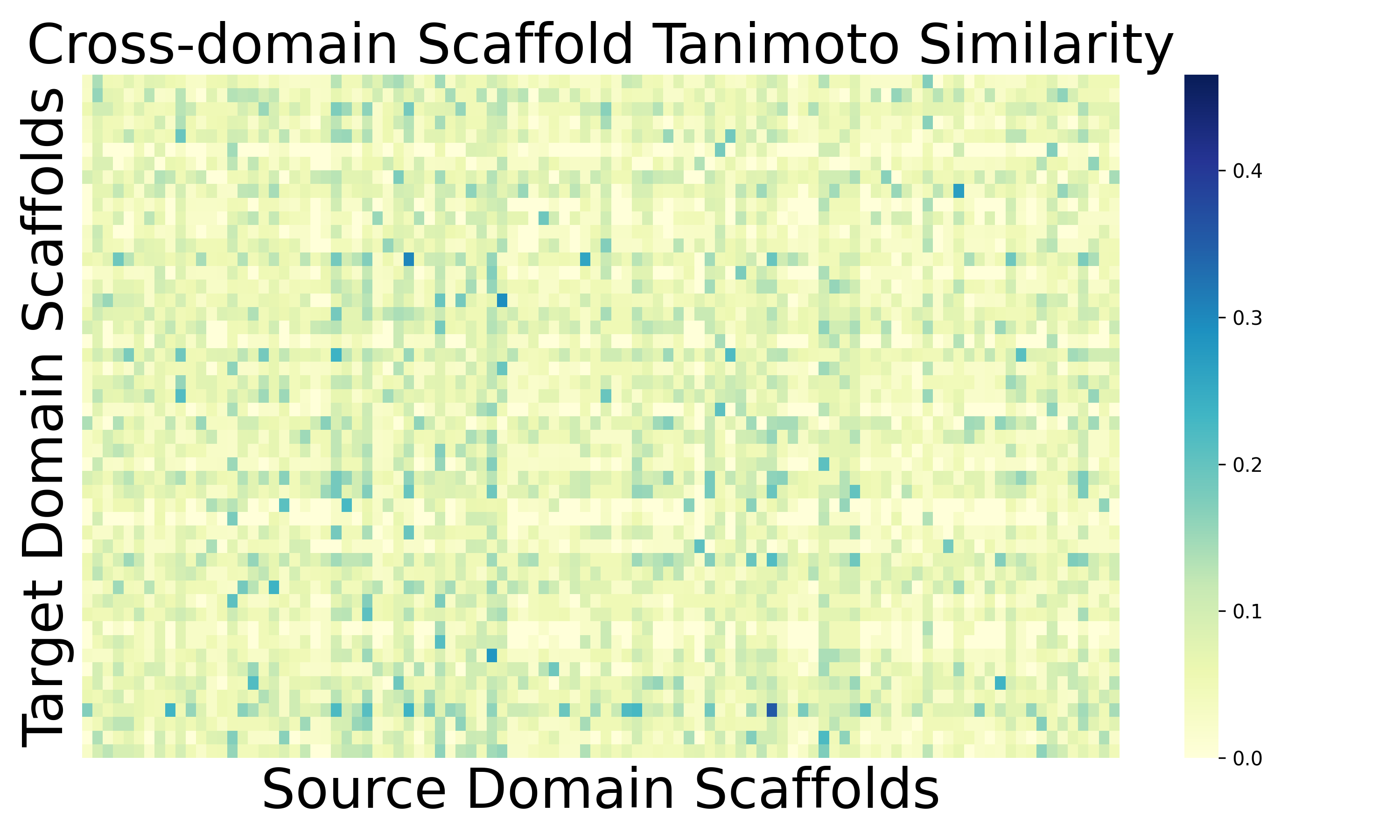}
    \caption{\bench{}}
  \end{subfigure}
  \caption{Inter-domain scaffold Tanimoto similarity heatmaps. Darker blue indicates higher similarity, whereas lighter yellow indicates lower similarity. The panels show qualitative differences in visible cross-domain similarity patterns across the three splitting protocols.}
  \label{fig:heatmap_compare}
\end{figure}

\paragraph{Scaffold similarity.} Figure~\ref{fig:heatmap_compare} provides the complementary qualitative comparison of inter-domain scaffold Tanimoto similarity under the three splitting protocols.

\FloatBarrier
\section{Detailed Source-Selection and Alignment Ablations}
\label{app:detailed_ablation}

Table~\ref{tab:detailed_ablation} compares source-selection heuristics and alignment configurations across 15 target scaffolds. The relative performance of the heuristic strategies varies by target; no heuristic is uniformly best across all scaffolds.

In the aggregate ViSNet--HOMO results, removing Mol-CORAL or Sub-CORAL increases MAE from 0.1540 to 0.1689 and 0.1672, respectively. The full model does not outperform both single-scale variants on every scaffold. For Scaffold 4, removing Sub-CORAL gives an MAE of 0.1473, compared with 0.1576 for the full model.

The full configuration has the lowest aggregate MAE in Table~\ref{tab:ablation}, while individual heuristic strategies obtain lower MAE on some scaffolds, including Scaffolds 7 and 15. These results show that the relative performance of selection strategies is target dependent.

\begin{table}[htbp]
  \centering
  \small
\caption{Detailed ablation study results measured in mean absolute error across fifteen individual target scaffolds for the HOMO property using the ViSNet architecture. The best performance for each target is highlighted in bold. The full framework is denoted as \ours{}.}
  \label{tab:detailed_ablation}
  \setlength{\tabcolsep}{3pt}
  \renewcommand{\arraystretch}{0.95}
  \resizebox{\textwidth}{!}{%
  \begin{tabular}{l c cccccc ccc}
    \toprule
    \multirow{2}{*}{Target} & \multirow{2}{*}{Baseline} & \multicolumn{6}{c}{Heuristic Source Retrieval Strategies} & \multicolumn{3}{c}{Module Ablation \& \ours{}} \\
    \cmidrule(lr){3-8} \cmidrule(lr){9-11}
    & & Random & Shallow & Deep & Physical & Graph-K. & Mixed & w/o Mol & w/o Sub & \ours{} \\
    \midrule
    Scaffold 1  & 0.1469 & 0.1458 & 0.1470 & 0.1347 & 0.1416 & 0.1364 & 0.1364 & 0.1328 & 0.1359 & \textbf{0.1302} \\
    Scaffold 2  & 0.1019 & 0.0870 & 0.0939 & 0.1003 & 0.0986 & 0.0897 & 0.0897 & 0.0982 & 0.0872 & \textbf{0.0835} \\
    Scaffold 3  & \textbf{0.1277} & 0.1618 & 0.1672 & 0.1562 & 0.1573 & 0.1422 & 0.1422 & \textbf{0.1277} & \textbf{0.1277} & \textbf{0.1277} \\
    Scaffold 4  & 0.2032 & 0.1857 & 0.1683 & 0.1784 & 0.1802 & 0.1628 & 0.1628 & 0.1697 & \textbf{0.1473} & 0.1576 \\
    Scaffold 5  & \textbf{0.1794} & 0.2001 & 0.2085 & 0.2029 & 0.1937 & 0.1878 & 0.1937 & \textbf{0.1794} & \textbf{0.1794} & \textbf{0.1794} \\
    Scaffold 6  & 0.1121 & 0.1095 & 0.1079 & 0.1060 & 0.1078 & 0.1005 & 0.1005 & 0.1026 & 0.1028 & \textbf{0.0926} \\
    Scaffold 7  & 0.2697 & 0.2950 & 0.2593 & 0.2755 & 0.2456 & \textbf{0.2386} & \textbf{0.2386} & 0.3171 & 0.3274 & 0.2584 \\
    Scaffold 8  & 0.0901 & 0.0886 & 0.0936 & 0.0960 & 0.0916 & \textbf{0.0871} & \textbf{0.0871} & 0.0941 & 0.0995 & 0.0888 \\
    Scaffold 9  & 0.2417 & 0.2326 & 0.2614 & \textbf{0.2299} & 0.2434 & 0.2336 & 0.2434 & 0.2417 & 0.2417 & 0.2417 \\
    Scaffold 10 & \textbf{0.1379} & 0.1744 & 0.1896 & 0.2101 & 0.1868 & 0.1972 & 0.1695 & \textbf{0.1379} & \textbf{0.1379} & \textbf{0.1379} \\
    Scaffold 11 & 0.1821 & 0.1920 & 0.1830 & 0.1948 & 0.1816 & 0.1853 & 0.1816 & 0.1821 & 0.1821 & \textbf{0.1768} \\
    Scaffold 12 & 0.1933 & 0.2042 & \textbf{0.1850} & 0.1941 & 0.1938 & 0.2131 & 0.1938 & 0.1933 & 0.1933 & 0.1933 \\
    Scaffold 13 & 0.2070 & 0.2435 & \textbf{0.1909} & 0.2161 & 0.1966 & 0.2216 & 0.1966 & 0.2070 & 0.2070 & 0.2070 \\
    Scaffold 14 & \textbf{0.0574} & 0.0636 & 0.0597 & 0.0734 & 0.0661 & 0.0635 & 0.0635 & 0.0800 & 0.0775 & 0.0751 \\
    Scaffold 15 & 0.1818 & 0.1774 & 0.1785 & \textbf{0.1349} & 0.1772 & 0.1652 & 0.1652 & 0.2700 & 0.2611 & 0.1594 \\
    \bottomrule
  \end{tabular}}
\end{table}

\section{Granular Sensitivity Analysis of Hyperparameters}
\label{app:hyperparameter_analysis}

Table~\ref{tab:detailed_hyperparameters} reports MAE for each of the 15 target scaffolds under the evaluated hyperparameter configurations, supplementing the aggregate results in the main text.

The per-scaffold results vary across the evaluated candidate-pool sizes. The default $M=50$ setting has MAE lower than or comparable to that of $M=20$ and $M=100$ on most reported scaffolds. On Scaffold 4, $M=20$ gives lower MAE.

Performance also varies with subset size. A subset of three sources is competitive on some target scaffolds, whereas the default configuration has the lowest prediction error on many evaluated scaffolds. The default setting is retained for the main experiments.

\begin{table}[htbp]
  \centering
  \small
\caption{Detailed hyperparameter sensitivity analysis measured in mean absolute error across fifteen individual target scaffolds for the HOMO property using the ViSNet architecture. The best performance for each target is highlighted in bold. The default configuration used in the main experiments is shown in the final column.}
  \label{tab:detailed_hyperparameters}
\setlength{\tabcolsep}{3pt}
\renewcommand{\arraystretch}{0.95}
  \begin{tabular}{l cc cc c}
    \toprule
    \multirow{2}{*}{Target}
      & \multicolumn{2}{c}{Candidate Pool Size $M$}
      & \multicolumn{2}{c}{Subset Size $K$}
      & \multirow{2}{*}{Default \ours{}} \\
    \cmidrule(lr){2-3} \cmidrule(lr){4-5}
      & $M=20$ & $M=100$ & $K=1$ & $K=3$ & $M=50$ and $K=5$ \\
    \midrule
    Scaffold 1  & 0.1376 & 0.1444 & 0.1492 & 0.1421 & \textbf{0.1302} \\
    Scaffold 2  & 0.0963 & 0.1086 & 0.0970 & 0.1078 & \textbf{0.0835} \\
    Scaffold 3  & \textbf{0.1277} & \textbf{0.1277} & \textbf{0.1277} & \textbf{0.1277} & \textbf{0.1277} \\
    Scaffold 4  & \textbf{0.1401} & 0.1852 & 0.1562 & 0.1941 & 0.1576 \\
    Scaffold 5  & \textbf{0.1794} & \textbf{0.1794} & \textbf{0.1794} & \textbf{0.1794} & \textbf{0.1794} \\
    Scaffold 6  & 0.1182 & 0.1239 & 0.1186 & 0.1060 & \textbf{0.0926} \\
    Scaffold 7  & 0.3004 & 0.2883 & 0.2939 & \textbf{0.2272} & 0.2584 \\
    Scaffold 8  & 0.0933 & 0.0940 & 0.0916 & 0.0980 & \textbf{0.0888} \\
    Scaffold 9  & \textbf{0.2417} & \textbf{0.2417} & \textbf{0.2417} & \textbf{0.2417} & \textbf{0.2417} \\
    Scaffold 10 & \textbf{0.1379} & \textbf{0.1379} & \textbf{0.1379} & \textbf{0.1379} & \textbf{0.1379} \\
    Scaffold 11 & 0.1821 & 0.1821 & 0.1821 & 0.1821 & \textbf{0.1768} \\
    Scaffold 12 & \textbf{0.1933} & \textbf{0.1933} & \textbf{0.1933} & \textbf{0.1933} & \textbf{0.1933} \\
    Scaffold 13 & \textbf{0.2070} & \textbf{0.2070} & \textbf{0.2070} & \textbf{0.2070} & \textbf{0.2070} \\
    Scaffold 14 & 0.0778 & 0.0804 & 0.0762 & 0.0754 & \textbf{0.0751} \\
    Scaffold 15 & 0.1869 & 0.2203 & 0.2569 & 0.3141 & \textbf{0.1594} \\
    \bottomrule
  \end{tabular}
\end{table}

\FloatBarrier
\section{Hyperparameter Settings}
\label{app:hyperparams}

Table~\ref{tab:hyperparams} summarizes all key hyperparameters used across experiments.

\begin{table}[H]
\centering
\small
\caption{Hyperparameter settings for \ours{}.}
\label{tab:hyperparams}
\setlength{\tabcolsep}{4pt}
\renewcommand{\arraystretch}{0.95}
\begin{tabular}{lll}
\toprule
Hyperparameter & Symbol & Value \\
\midrule
Candidate pool size & $M$ & 50 \\
Proxy targets per policy-training stage & $N_p$ & 20 \\
Maximum training subset size & $K_{\max}$ & 5 \\
Deployment subset size & $K_{\mathrm{deploy}}$ & 5 \\
GRPO group size (including empty control) & $G$ & 33 \\
GRPO training steps & & 40 \\
KL penalty coefficient & $\beta$ & 0.01 \\
Clip threshold & $\epsilon_{clip}$ & 0.2 \\
HubScore threshold & $\tau_{sim}$ & 0.45 \\
HubScore balance coefficient & $\lambda$ & 0.3 \\
Warm-up epochs & $E_{warm}$ & 10 \\
Regression threshold & $\tau_{reg}$ & 0.05 \\
Momentum coefficient & $\beta_m$ & 0.9 \\
Numerical stability & $\epsilon_s$ & $10^{-8}$ \\
Backbone learning rate & & $3\times10^{-4}$ \\
Batch size & & 64 \\
\bottomrule
\end{tabular}
\end{table}

\end{document}